\documentclass[sigconf]{acmart}

\AtBeginDocument{%
  \providecommand\BibTeX{{%
    \normalfont B\kern-0.5em{\scshape i\kern-0.25em b}\kern-0.8em\TeX}}}

\setcopyright{acmlicensed}
\copyrightyear{2018}
\acmYear{2018}
\acmDOI{XXXXXXX.XXXXXXX}

\acmConference[MM'24]{Make sure to enter the correct
  conference title from your rights confirmation email}{October 28 - November 1,
  2024}{Melbourne, Australia.}
\acmISBN{978-1-4503-XXXX-X/18/06}

\usepackage{multirow}
\usepackage{mathrsfs}
\usepackage{amsmath}
\usepackage{bm}
\usepackage{graphicx}
\usepackage{multirow}
\usepackage{tabularray}
\usepackage{booktabs}
\usepackage{makecell}
\usepackage{amssymb}

\begin{document}

\title{Revisiting the Robust Generalization of Adversarial Prompt Tuning}


\author{Fan Yang, Mingxuan Xia, Sangzhou Xia, Chicheng Ma, Hui Hui}
\renewcommand{\shortauthors}{author name and author name, et al.}

\begin{abstract}
Understanding the vulnerability of large-scale pre-trained vision-language models like CLIP against adversarial attacks is key to ensuring zero-shot generalization capacity on various downstream tasks. State-of-the-art defense mechanisms generally adopt prompt learning strategies for adversarial fine-tuning to improve the adversarial robustness of the pre-trained model while keeping the efficiency of adapting to downstream tasks.
Such a setup leads to the problem of over-fitting which impedes further improvement of the model's generalization capacity on both clean and adversarial examples.
In this work, we propose an adaptive Consistency-guided Adversarial Prompt Tuning (i.e., CAPT) framework that utilizes multi-modal prompt learning to enhance the alignment of image and text features for adversarial examples and leverage the strong generalization of pre-trained CLIP to guide the model-enhancing its robust generalization on adversarial examples while maintaining its accuracy on clean ones. We also design a novel adaptive consistency objective function to balance the consistency of adversarial inputs and clean inputs between the fine-tuning model and the pre-trained model. We conduct extensive experiments across 14 datasets and 4 data sparsity schemes (from 1-shot to full training data settings) to show the superiority of CAPT over other state-of-the-art adaption methods. CAPT demonstrated excellent performance in terms of the in-distribution performance and the generalization under input distribution shift and across datasets.
\end{abstract}





\maketitle

\section{Introduction}
Large-scale models pre-trained on vision and language data have emerged as vision-language foundation models \cite{radford2021learning,jia2021scaling,li2022blip} achieving great success in numerous fields such as visual question answering \cite{zhou2020unified, lin2022retrieval}, image captioning \cite{zhu2023prompt}, text-to-image generation \cite{li2023gligen, zhang2023adding}. With the rapid development of the multi-modal foundation model, various Visual-Language models (VLMs) have been proposed, and an increasing number of studies have introduced VLMs into different downstream tasks.

Unfortunately, a lot of  recent studies \cite{mao2022understanding,schlarmann2023adversarial,zhao2024evaluating} unveiled that VLMs are vulnerable to small adversarial noise \cite{szegedy2013intriguing}---the model will output completely different error results by introducing deliberately designed imperceptible perturbations. 
Due to the fact that VLMs become susceptible to adversarial example attacks when applied to downstream tasks, it is critical to increase the model's robustness to ensure its dependable use in downstream tasks.
In order to mitigate the risk posed by adversarial examples and enhance the robustness of models against adversarial attacks, numerous defense strategies have been suggested. 
Among them, adversarial training \cite{madry2018towards,zhang2019theoretically,wang2019improving} is one of the most common and effective approaches for adversarial defense which can be regarded as a type of data augmentation technique that crafts adversarial versions of the natural examples for model training. 
As for large-scale models, applying adversarial training from scratch to improve its robustness is impractical due to the computation-consuming process of adversarial examples generation in each training step, especially in models with massive parameters. 
Fine-tuning \cite{yosinski2014transferable} is a relatively efficient way to better adapt pre-trained models to downstream tasks, however, as pre-trained models expand to encompass tens or hundreds of billions of parameters, the process of fine-tuning all these model weights becomes exceedingly expensive. This cost can escalate further if adversarial training is employed to enhance the robustness of the model against adversarial attacks. 

To this end, prompt tuning \cite{zhou2022learning,zhou2022conditional,lu2022prompt,khattak2023maple,yao2023visual,jia2022visual} is applied as a more efficient alternative fine-tuning approach to fine-tuning which enables the model to transfer to downstream tasks.  
Without changing the pre-trained models' weights, this method adapts the pre-trained model to downstream tasks by adding some learnable prompt vectors. 
When using adversarial training to fine-tune downstream tasks for large-scale pre-trained models, adopting prompt learning can improve the model's robustness while improving training efficiency. 
Recently, several works employed this efficient approach to enhance the adversarial robustness of pre-trained models in downstream tasks\cite{chen2020adversarial,huang2023improving, li2024one}. Huang \cite{huang2023improving}  and Chen et al. \cite{chen2023visual} have investigated the use of adversarial visual prompting \cite{bahng2022exploring} as a means of defense at test time to improve the adversarial robustness of pre-trained models. Li et al. \cite{li2024one} found that the adversarial robustness of CLIP is sensitive to the prompt used for inference. So they adopt text prompt tuning to adapt pre-trained models. 

However, these methods suffer a few drawbacks. All these approaches only focus on the adversarial examples but ignore their clean counterpart. Moreover, none of them pay attention to the problem of over-fitting in adversarial prompt learning \cite{salman2020adversarially,rice2020overfitting, zhou2022conditional,khattak2023self}. Under these circumstances, the model is influenced by the distribution of adversarial examples on small-scale training set, which will result in a decrease of the model's robust generalization on both adversarial examples and clean ones.

In this work, we move one step and propose an adaptive Consistency-guided Adversarial Prompt tuning (CAPT) which utilizes multi-modal prompt learning and the powerful generalization ability of the pre-trained CLIP to improve the adversarial robust generalization of the pre-trained model while maintaining its accuracy on clean examples. Inspired by \cite{khattak2023maple,khattak2023self}, we adopt multi-modal prompt learning to improve the alignment between visual and textual features for adversarial examples and enhance the robustness of image and text encoder during training. Previous methods only utilize the adversarial examples for adversarial fine-tuning, inspired by TRADES \cite{zhang2019theoretically}, our approach focuses on the consistency between clean and adversarial inputs and adds a KL divergence regularization term to ensure this consistency. Moreover, we leverage the generalization ability of frozen pre-trained CLIP to tackle the over-fitting issue of adversarial fine-tuning and improve the zero-shot adversarial robustness. A regularization loss is also introduced to improve the model’s adversarial robust generalization capabilities. To balance the consistency between clean and adversarial inputs of the fine-tuned model and frozen pre-trained model, We design a novel adaptive consistency loss function that dynamically adjusts the weight of the different losses based on the model's reliability. This weight determines the balance between learning from a pre-trained model and relying on fine-tuning the model's own. 
Following APT \cite{li2024one}, we conduct extensive experiments on 14 datasets and 4 data sparsity schemes, 1-, 4- and
 16-shot learning and training with the entire training set.
Our contributions are summarized as follows:
\begin{itemize}
\item We investigate the over-fitting problem in adversarial prompt tuning and discuss the drawback of adversarial prompt tuning as well as reason causing the over-fitting issue.
\item We propose a novel adaptive Consistency-guided Adversarial Prompt Tuning approach, which introduces powerful multi-modal prompts and a novel adaptive consistency-guided objective function leveraging the generalization capacity of pre-trained frozen CLIP. Our approach achieves significant improvement in robust generalization of the pre-trained model.
\item We conduct an extensive evaluation on different datasets including in-distribution experiments and out-of-distribution experiments. The results demonstrate that our proposed method significantly outperforms all other state-of-the-art baselines indicating the superiority of our method.
\end{itemize}
To the best of our knowledge, we are the first to investigate the over-fitting problem of adversarial prompt tuning. Specifically, our method significantly outperforms the state-of-the-art by 16.91\% and 7.51 \% for accuracy and robustness respectively averaged on different shots on the ImageNet dataset. ~\footnote{The code is available in the supplementary materials.}.

\section{Related Works}

\textbf{Vision Language models} 
Foundational vision-language models(VLMs) \cite{radford2021learning,jia2021scaling,li2022blip,li2023blip} utilize both visual and textual data to learn rich semantic multi-modal representations. These models are pre-trained on large-scale multi-modal datasets through image-text contrastive learning, which effectively draws the features of corresponding image-text pairs closer together, while simultaneously pushing away the features of mismatched pairs. By leveraging large-scale image-text datasets, for instance, 400 million pairs for CLIP \cite{radford2021learning} and 1 billion for ALIGN \cite{jia2021scaling}, and employing end-to-end pre-training strategies, VLMs can learn rich semantic associations between images and text enable them to better understanding multi-modal information of open vocabulary concept. Utilizing the powerful generalization ability of the pre-trained model, VLMs achieve state-of-the-art performance on various visual and vision-language tasks \cite{bangalath2022bridging,maaz2022class,gu2021open,zang2022open,li2022language,rao2022denseclip}.

\textbf{Prompt Tuning for VLMs} 
Prompt tuning \cite{zhou2022learning,zhou2022conditional,khattak2023maple,gao2024clip,zhu2023prompt,yao2023visual} has been introduced as an effective fine-tuning strategy to adapt pre-trained VLMs to specific downstream tasks. This approach incorporates a few trainable embeddings along with model inputs which are optimized during training while the rest of the model is kept frozen. Since the pre-trained model remains unchanged during the prompt learning, this strategy has proven to be especially beneficial for Visual Language Models (VLMs) like CLIP, where preserving the model's inherent ability to generalize is essential. Context Optimization(CoOp) \cite{zhou2022learning} replaced the hand-crafted prompts with learnable textual prompts to improve the textual embedding. Conditional Context Optimization(CoCoOp) \cite{zhou2022conditional} focuses on the overfitting problem of CoOp and proposes to condition prompts based on visual features for improved performance on generalization tasks. In addition, Knowledge-Guided Context Optimization(KgCoOp) \cite{yao2023visual} ensures that the designed learnable prompts incorporate crucial general knowledge. PLOT \cite{chen2022plot} utilizes optimal transport to align the vision and text modalities, creating discriminative and visually coherent local textual prompts. Beyond textual prompt tuning, Multi-modal Prompt Learning (MaPLe) \cite{khattak2023maple} and PromptSRC \cite{khattak2023self} enhance the process by tuning prompts across both visual and text encoders simultaneously.

\textbf{Adversarial Robustness}  
Deep neural networks are susceptible to adversarial attacks, where barely noticeable noises are added to original images, causing incorrect classifications by the models \cite{goodfellow2014explaining,madry2017towards,dong2018boosting}. To counteract this vulnerability, various defense strategies have been devised. Notably, adversarial training \cite{jia2022adversarial,pang2022robustness,wang2019improving,zhang2019theoretically} stands out as an effective defense method. This technique incorporates adversarial examples into the training dataset during the model's training phase, significantly bolstering the DNNs' resistance to such attacks. As the adoption of large-scale pre-trained vision language models increases, their susceptibility to adversarial threats has also come into focus, with a proliferation of attack algorithms specifically designed against them \cite{yin2023vlattack,zhang2022towards,lu2023set,zhao2024evaluating}. Recently, several studies have explored enhancing the adversarial robustness of pre-trained models through adversarial fine-tuning, which involves adjusting the model's weights via adversarial training. Some of them improve the robustness of the model by fine-tuning the entire model\cite{mao2022understanding,li2023language,wang2024pre}. Mao et al. \cite{mao2022understanding} proposed a text-guided contrastive adversarial training loss and applied it for adversarial fine-tuning in order to boost the zero-shot adversarial robustness of large-scale pre-trained models. Wang et al. \cite{wang2024pre} introduced a pre-trained model-guided adversarial fine-tuning to enhance the model's zero-shot adversarial robustness. This approach fine-tuned the whole image encoder of the pre-trained model, which is inefficient and may destroy the generalization of the pre-training parameters. A few other methods employ partial adversarial fine-tuning \cite{chen2023visual, chen2020adversarial, huang2023improving, li2024one}. Huang \cite{huang2023improving}  and Chen et al. \cite{chen2023visual} have investigated the use of adversarial visual prompting \cite{bahng2022exploring} as a means of defense at test time to improve the adversarial robustness of pre-trained models. Li et al. \cite{li2024one} found that the adversarial robustness of CLIP is sensitive to the prompt used for inference. So they adopt text prompt tuning to adapt pre-trained models.

Although these methods can improve the robustness of pre-trained models, they ignore the over-fitting problem of adversarial fine-tuning \cite{dodge2020fine,rice2020overfitting} especially for prompt learning \cite{zhou2022conditional,khattak2023self}.  
Different from them, we propose an adaptive Consistency-guided Adversarial Prompt tuning (CAPT) which utilizes multi-modal prompt learning \cite{khattak2023maple} and the powerful generalization ability of the pre-trained CLIP to improve the adversarial robust generalization of the pre-trained model while maintaining its accuracy on clean examples. 

\section{Proposed Method}
We first give preliminary in Sec. 3.1 on CLIP and Adversarial Prompt Tuning. In Sec. 3.2, we provide the details of our 
adaptive Consistency-guided Adversarial Prompt Tuning (CAPT) framework including the main component of our framework and our novel objective function.

\subsection{Preliminaries}
\textbf{CLIP Revisiting} 
CLIP model is mainly composed of two parts: image encoder and text encoder. We denote the image encoder and text encoder as $f_{\mathcal{I}}$ and $f_{\mathcal{T}}$ respectively and their pre-trained parameters as $\theta_{\mathcal{I}}$ and $\theta_{\mathcal{T}}$. The deep features of images and text can be extracted by the corresponding encoder respectively. For the visual branch, the input image $\boldsymbol{x}_i$ is firstly divided into $M$ patches and project to patch embedding. After adding a learnable class token $\boldsymbol{e}_{cls}$, the input embedding is encoded by the image encoder to produce a latent visual representation $\boldsymbol{z}_v^i=f_{\mathcal{I}}(\boldsymbol{x}_i,\theta_{\mathcal{I}})$, where $\boldsymbol{z}_v^i$ is a $d$ dimensional feature vector for the image i. For the text branch, the class label $y$ is wrapped within a text template for example a photo of a \{class label\}’. Through word embedding, we can get the input text features of class j $\boldsymbol{t}_j$ for the text encoder. The text encoder encodes $\boldsymbol{t}_j$ via stacks of transformer blocks to produce a latent textual feature $\boldsymbol{z}_t^j=f_{\mathcal{T}}(\boldsymbol{t}_j,\theta_{\mathcal{T}})$. After getting the latent image feature $\boldsymbol{z}_v^i$ and the latent textual feature $\boldsymbol{z}_t^j$, we could perform zero-shot inference by calculating the probability of class j for the image $i$ as
\begin{equation}
  p_{i,j} = \frac{{\rm exp(cos}(\boldsymbol{z}_v^i ,\boldsymbol{z}_t^j)/\tau)}{\sum_{j=1}^{C}{\rm exp(cos}(\boldsymbol{z}_v^i ,\boldsymbol{z}_t^j)/\tau)}
\end{equation}

\textbf{Adversarial Prompt Tuning} Adversarial attacks are commonly conducted by generating an imperceptible perturbation for clean input misleading the model to produce a wrong prediction. For CLIP is to search for a perturbation $\boldsymbol{\delta}_i$ for input $\boldsymbol{x}_i$ to maximize the dissimilarity between the image feature $\boldsymbol{z}_v^i$ and the text feature of the ground-truth class prompt, $\boldsymbol{z}_t^{y_{i}}$. Assuming $\boldsymbol{\delta}$ is bounded by $\epsilon$-ball of $p$-norm, which can be formulated as:
\begin{equation}
\mathop{\arg\max}\limits_{\Vert \boldsymbol{\delta}_{i} \Vert \leq \epsilon } L(\boldsymbol{x}_i + \boldsymbol{\delta}_i, \boldsymbol{t}_{y_j}, y_{j};\boldsymbol{\theta}_{\mathcal{I}},\boldsymbol{\theta}_{\mathcal{T}})
\end{equation}
Li et al. \cite{li2024one} discover that the robustness will change a lot when using different prompts for inference, which indicates that the adversarial robustness of VLMs is sensitive to the text prompt during inference. Besides, the hand-crafted text prompt contains no additional information about the adversarial samples. To address this problem, they proposed to improve the adversarial robustness of VLMs through adversarial prompt tuning(APT). Following CoOp\cite{zhou2022learning}, they replace the word embedding of a fixed prompt template with a photo of a \{class label\}’ to sequence of $M$ learnable vector with a class embedding $[C_j]$ as 
\begin{equation}
\boldsymbol{t}_j=[V]_1[V]_2...[V]_m[C_j]
\end{equation}
To improve adversarial robustness, they adopt adversarial training during prompt tuning. Adversarial training can be formulated as a min-max optimization problem which can be described as:
\begin{equation}
\mathop{\min}\limits_{\boldsymbol{t}} \mathop{\max}\limits_{\Vert \boldsymbol{\delta}_{i} \Vert \leq \epsilon } L(\boldsymbol{x}_i + \boldsymbol{\delta}_i, \boldsymbol{t}, y_{j};\boldsymbol{\theta}_{\mathcal{I}},\boldsymbol{\theta}_{\mathcal{T}})
\end{equation}
In order to get the model to make the right predictions on adversarial instances, the inner maximizing process creates examples that are deceptive to the model's prediction, while the outer minimization process optimizes the learnable prompt vector. The adversarial resilience of the pre-trained model was steadily strengthened when the optimization processes of maximizing and minimization alternated. Nevertheless, the clean inputs and the association between adversarial and clean inputs are disregarded by Adversarial Prompt Tuning. The over-fitting issue in adversarial training and quick tweaking is also of little consequence to APT. 

\begin{figure*}[!t]
\centerline{\includegraphics[width=0.88\textwidth]{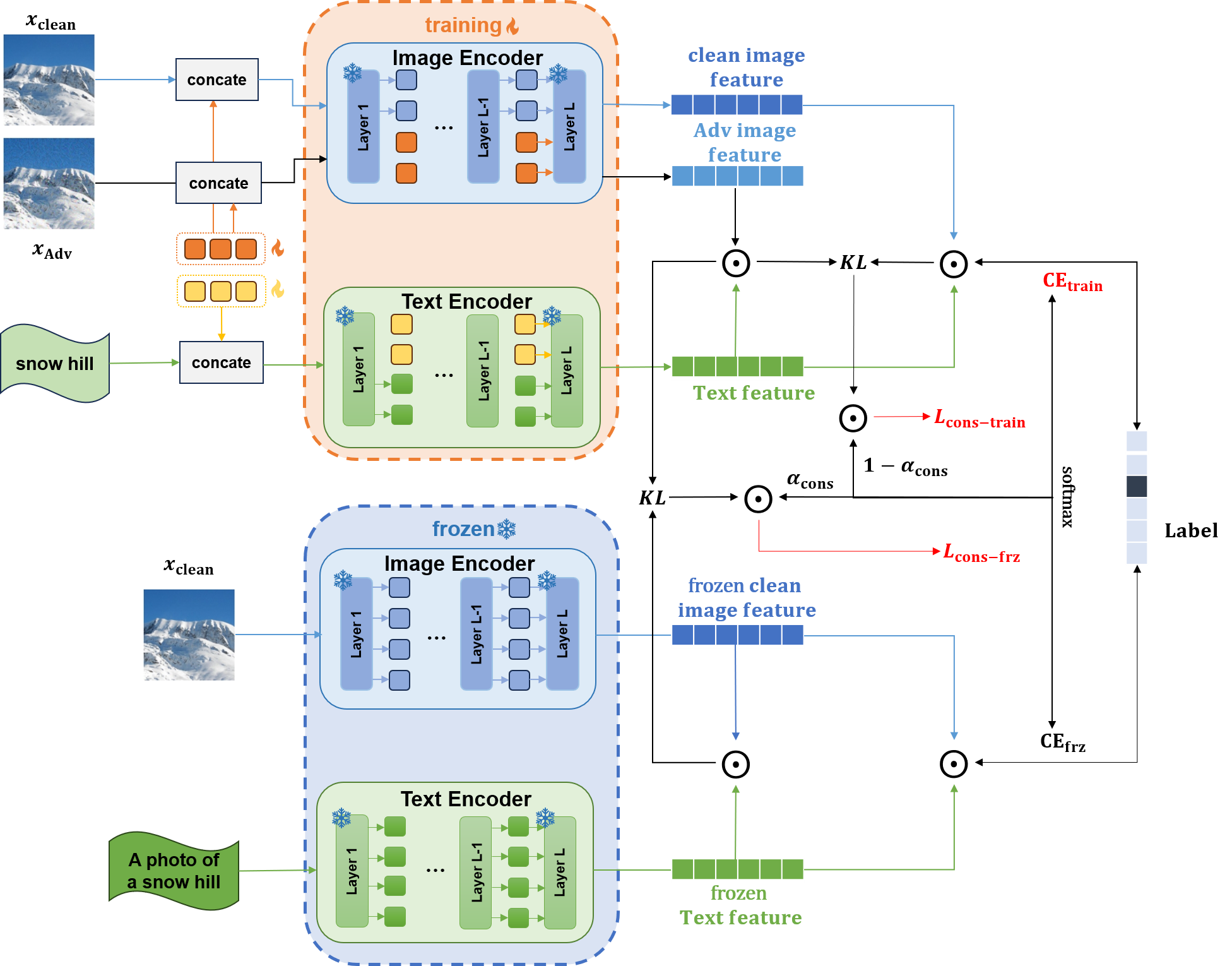}}
\caption{The overview of our adaptive Consistency-guided Adversarial Prompt Tuning framework. Our method adopts multi-modal prompt learning to improve the alignment between visual and textual features for adversarial examples and enhance the robustness of image and text encoder during training. we introduce a frozen pre-trained CLIP to tackle with the over-fitting issue of adversarial fine-tuning and improve the zero-shot adversarial robustness.
}
\label{fig1arch}
\end{figure*}

\subsection{CAPT}
To efficiently improve the robustness of CLIP  when adapting to downstream tasks, several researchers adopt prompt tuning strategies in adversarial fine-tuning. Chen et al. \cite{chen2023visual} attempt to utilize adversarial visual prompting to improve the adversarial robustness in test time. Li et al. \cite{li2024one} found that a slight change of prompt will influence the robustness which indicates that the adversarial robustness is sensitive to the choice of prompt for inference. Thus, they use learnable textual prompts instead of fixed prompts to improve the adversarial robustness. However, all the existing adversarial prompt tuning methods follow uni-modal solutions either in the vision or in the language branch of CLIP. They ignore that both image and text encoder contribute to the alignment of multi-modal features especially for the adversarial examples whose feature distribution is quite different from the clean ones. In adversarial situations, learning prompts only for text encoders are insufficient for accurate prediction. Therefore, in order to improve the alignment of multi-modal features of adversarial examples and enhance the robustness of both image encoder and text encoder, we applied multi-modal prompt tuning \cite{khattak2023maple} which employ learnable prompt vector in both image encoder and text encoder of CLIP. Following khattak \cite{khattak2023maple}, we learn prompts in the deeper transformer layers to progressively model stage-wise feature representations. We introduce b learnable tokens $V_l=\{v_l^1,v_l^2,...,v_l^b\}$ and $T_l=\{T_l^1,T_l^2,...,T_l^b\}$ respectively as the learnable vision and text prompts in the $l_th$ transformer layer. New learnable prompts are introduced in each transformer block of image and text encoder up to depth $J$. For the visual branch, the process can be described as,
\begin{align}
[\_, E_i] = f_{\mathcal{I}}^i(V_{i-1},E_{i-1}), \quad i = 1,2,...,J \\
[V_i, E_i] = f_{\mathcal{I}}^i(V_{i-1},E_{i-1}), \quad i = J+1,...,K
\end{align}
$[]$ is the concatenation operation. $f_{\mathcal{I}}^i$ is the process of image encoder in the layer $i$. $E_i$ is the output image feature in the layer $i$.

For the text branch, it is similar to the visual branch, which can be described as,
\begin{align}
[\_, W_j] = f_{\mathcal{T}}^j(T_{j-1},W_{j-1}), \quad j = 1,2,...,J \\
[T_j, W_j] = f_{\mathcal{T}}^j(T_{j-1},W_{j-1}), \quad j = J+1,...,K
\end{align}
$f_{\mathcal{T}}^j$ is the process of text encoder in the layer $j$. $W_j$ is the output text feature in the layer $j$. Applying multi-modal prompting in the deep layer enables learning prompts across various feature hierarchies within the transformer architecture.

\textbf{Trade-off generalization between robustness and accuracy} 
Adversarial prompt tuning and other approaches adopt adversarial training \cite{madry2018towards} strategy to improve the adversarial robustness of the pre-trained model when adapting to downstream tasks. However, all these methods only focus on the discrepancy between the prediction of adversarial inputs and the ground-truth labels but ignore clear examples. Although the adversarial robustness of the model has been improved during this process, the model's generalization capacity on clean examples will potentially decrease. During the adaptation of pre-trained models to downstream tasks, it is very important to enhance the model's adversarial robustness while ensuring accurate predictions for clean samples. Inspired by TRADES\cite{zhang2019theoretically} in adversarial training, we consider both the accuracy and robustness of the pre-trained model on clean examples and adversarial examples. Besides, we leverage the correlation between the clean examples and adversarial examples to improve robustness, the objective function can be described as,
\begin{equation} 
\begin{split}
\label{eq_trades}
    L_{\rm trades} &={\rm CE}({\rm sft}({\boldsymbol{z}}^I \cdot \boldsymbol{z}^T / \tau), \boldsymbol{y})  \\ 
    & + \lambda {\rm KL}({\rm sft} (\boldsymbol{z}^{I_a} \cdot \boldsymbol{z}^T / \tau),{\rm sft}(\boldsymbol{z}^I \cdot \boldsymbol{z}^T / \tau))
\end{split}
\end{equation}
where $\boldsymbol{z}^I$ and $\boldsymbol{z}^{I_a}$ represent the latent visual features of clean inputs and adversarial inputs respectively. $\boldsymbol{z}^T$ is the latent text feature for the class labels. $\boldsymbol{y}$ is the ground-truth label of the inputs. sft stands for the softmax operation. CE denotes the cross-entropy and KL denotes the Kullback–Leibler divergence. $\lambda$ is a weight balancing the focus on accuracy or robustness during adversarial fine-tuning.

In such an objective function, with the assistance of multi-modal prompts, pre-trained models can accurately predict clean samples while adapting to downstream tasks and narrow the distance between adversarial and clean samples, thereby enhancing the model's adversarial robustness. This achieves a balance in the model's generalization capabilities across both clean and adversarial samples.

\textbf{Adaptive consistency guided with the frozen CLIP} 
Prompt tuning is an efficient strategy for pre-trained models to adapt and learn task-specific knowledge without modifying the original pre-trained parameters of the model. During training, learnable prompts interact with frozen CLIP tokens via self-attention within the transformer architecture. This interaction of prompt tokens with pre-trained CLIP parameters could retain the generalization capacity of the pre-trained model within learned prompts implicitly. However, it still suffers from the problem \cite{zhou2022conditional,yao2023visual,khattak2023self} of over-fitting when fine-tuning on specific downstream tasks, which leads to a degeneration of the pre-trained model's robust generalization capacity. Within the adversarial training framework, the over-fitting problem will be further aggravated \cite{rice2020overfitting,wang2024balance}as the model is influenced by the distribution of generated adversarial examples of the training set which leads to a decline of robustness of adversarial examples on unseen data. Thus in order to enhance the robust generalization of the pre-trained model on the downstream tasks, inspired by \cite{khattak2023self}, we introduce a pre-trained frozen CLIP model with great generalization capacity to explicitly guide the adversarial features from the prompt tuning model to be consistent with the clean features from the frozen CLIP, which can be described as
\begin{equation}
    L_{\rm cons-frz} = {\rm KL}(({\rm sft} (\boldsymbol{z}^{I_a} \cdot \boldsymbol{z}^T / \tau),{\rm sft}(\boldsymbol{z}_{\rm frz}^I \cdot \boldsymbol{z}_{\rm frz}^T / \tau))
\end{equation}
$\boldsymbol{z}_{frz}^I$ and $\boldsymbol{z}_{frz}^T$ are the image latent features and text latent features for the clean examples from the pre-trained frozen CLIP.

With the assistance of pre-trained frozen CLIP, our approach could learn more generalized features for adversarial examples in order to improve the robust generalization of the model during adversarial fine-tuning.

In addition, the second term of the origin TRADES-like objective function in equation \ref{eq_trades} guides the prompt to focus on the correlation between adversarial examples and clean examples in the prompt-tuning model which learns task-specific features when adapting to downstream tasks. Here, we rewrite as 
\begin{equation}
    L_{\rm cons-train} = {\rm KL}(({\rm sft} (\boldsymbol{z}^{I_a} \cdot \boldsymbol{z}^T / \tau),{\rm sft}(\boldsymbol{z}^I \cdot \boldsymbol{z}^T / \tau)))
\end{equation}

In order to guide prompts to balance the consistency between the pre-trained frozen CLIP and the adversarial prompt tuning CLIP itself, we design an adaptive weighting strategy in stead of using fixed weight parameter for these two terms. Specifically, we assign different weights to different losses based on the reliability of the model's predictions on clean examples. The specific formulation of the adaptive consistency guided loss is as follows:
\begin{gather}
\label{consistency}
    L_{\rm adv-cons}  =  (1-\alpha_{\rm cons})L_{\rm cons-train} + \alpha_{\rm cons} L_{\rm cons-frz}  \\
    \alpha_{\rm cons}  = \frac{{\rm exp}({\rm CE}({\rm sft}({\boldsymbol{z}}_{\rm frz}^I \cdot {\boldsymbol{z}}_{\rm frz}^T / \tau), \boldsymbol{y}))}{{\rm exp}({\rm CE}({\rm sft}({\boldsymbol{z}}_{\rm frz}^I \cdot \boldsymbol{z}_{\rm frz}^T / \tau), \boldsymbol{y}) + {\rm exp}({\rm CE}({\rm sft}({\boldsymbol{z}}^I \cdot \boldsymbol{z}^T / \tau), \boldsymbol{y})} \notag
\end{gather}
where ${\alpha_{\rm cons}}$ is the adaptive weight parameter which is calculated according to the reliability of the frozen CLIP. Adaptive weights can dynamically adjust during the training process, guiding the prompt to balance the consistency between similarity for the features of adversarial examples and the features of clean examples obtained from the frozen CLIP and the fine-tuning CLIP. Combining this with Cross-Entropy (CE) loss enables the adversarial prompt tuning process to adapt to downstream tasks, maximizing model performance while ensuring that the learned features are consistent with the generalizable features of the frozen CLIP, thereby enhancing the model's robust generalization.

Finally, we got the overall objective function of our adaptive consistency adversarial prompt tuning (CAPT) framework:

\begin{align}
    L_{\rm CAPT} & =  {\rm CE}({\rm sft}({\boldsymbol{z}}^I \cdot \boldsymbol{z}^T / \tau), \boldsymbol{y}) + \lambda L_{\rm adv-cons} \\
    & =  {\rm CE}({\rm sft}({\boldsymbol{z}}^I \cdot \boldsymbol{z}^T / \tau), \boldsymbol{y}) \notag \\
    & +  \lambda[(1-\alpha_{\rm cons})L_{\rm cons-train} + \alpha_{\rm cons} L_{\rm cons-frz}] \notag
\end{align}

CAPT focuses on both the clean examples and adversarial examples during training. In addition, with the assistance of adaptive weighting, CAPT achieves a balance between the frozen CLIP and the fine-tuning model of the alignment of the adversarial examples and their counterparts.
\section{Experiments}
In this section, we conduct an extensive evaluation on different datasets including in-distribution experiments and out-of-distribution experiments. The results demonstrate that our proposed method significantly outperforms all other state-of-the-art baselines indicating the superiority of our method.\\
\textbf{Datasets.} 
The experiments in this section were the same as the Adversarial Prompt Tuning setup unless otherwise specified. Following Li et al. \cite{li2024one}, we use the same 11 datasets to evaluate our method for fairness: ImageNet\cite{deng2009imagenet}, Caltech101\cite{fei2004learning}, OxfordPets\cite{parkhi2012cats}, StanfordCars\cite{krause20133d}, Flowers102\cite{nilsback2008automated}, Food101\cite{bossard2014food}, FGVCAircraft\cite{maji2013fine}, SUN397\cite{xiao2010sun}, DTD \cite{cimpoi2014describing}, EuroSAT \cite{helber2019eurosat} and UCF101 \cite{soomro2012ucf101}. For each dataset, we evaluate with N-shots, meaning N examples per class are randomly sampled from the entire training set for training. N was either 1, 4, 16, or “all”, where "all" means the entire training set was used. One exception was for ImageNet, where 100 shots were used instead of “all” because our computational resource was insufficient to run experiments on the full dataset. All methods are evaluated on the entire test set regardless of the training data scheme used. 

\textbf{Baseline methods.} 
Our proposed method is a multi-modal prompting-based parameter-efficient adaption method. We compare it against two groups of related work: text prompting and prompting-based adversarial fine-tuning methods. For text prompt, we compare our method against Hand-Engineered Prompts (HEP) \cite{radford2021learning} which was originally proposed in CLIP and has been widely used in other works \cite{zhao2024evaluating,zhou2022learning,zhou2022conditional}. For prompting-based adversarial fine-tuning methods, we adopt Adversarial Visual Prompting(AVP) \cite{chen2023visual}, Partial Adversarial Fine-Tuning (PAFT) \cite{chen2020adversarial} and Adversarial Prompt Tuning (APT) \cite{li2024one} for comparison.  AVP utilizes adversarial visual prompting \cite{bahng2022exploring} as a test-time defense to enhance adversarial robustness for pre-trained models. PAFT can be viewed as the adversarial training variant of linear probing \cite{radford2021learning}. APT combines textual prompt tuning with adversarial training to improve the robustness of the pre-trained model when adapting to downstream tasks. All compared methods share the same frozen pre-trained image and text encoders.

\textbf{Implementation Details.} 
We utilize the ViT-B/32 architecture of the CLIP model as the backbone. Following APT, the weights of image encoders were pre-trained using the state-of-the-art zero-shot adversarial robustness method TeCoA \cite{mao2022understanding}. We use the SGD optimizer with a momentum of 0.9. The initial learning rate is set at 0.0025 with a cosine learning rate scheduler and a warm-up strategy during the first epoch. We use context length 16 and independent V-L prompting in the first 9 transformer layers for both image and text branches. For hyper-parameters, we empirically set $\lambda$ in the objective function to 100. The batch size is 32 for all the datasets. For ImageNet, the number of epochs was 20,20,50, and 20 for 1,4,16 and all shots respectively, and for other datasets was 50,100,200,200 which keeps the same with APT. Results were reported for the last checkpoint. The PGD \cite{madry2018towards} attack is used for both training and evaluation. Two perturbation budgets, $\epsilon=1/255$ and $4/255$ are used following \cite{mao2022understanding}
 and \cite{croce2020robustbench} respectively. We use 3 steps with a step size of $2\epsilon/3$ for training and 100 steps with a step size of $\epsilon/4$ and random start for evaluation~\footnote{More details of the experimental setup and results are in the supplementary materials.}. 

\section{Main Result}
\subsection{In-Distribution Performance}
We evaluate the performance of our method and other baseline methods on the in-distribution scenarios, where the training and test data share the same distribution. We conduct experiments on different shots (1,4,16, all) and different perturbations ($\epsilon=1/255$ and $4/255$). A comparison of different prompting methods on different datasets and shots for various perturbations is shown in Table \ref{tab:tab1}. We can see that compared with the hand-engineered prompts our method achieves a significant improvement in both accuracy and robustness. Besides, with the increase in the number of shots, the improvement is further increased. Compared to other prompting-based adversarial fine-tuning methods, our method also achieves huge superiority against all other methods on accuracy and robustness, especially for the $\epsilon=4/255$ perturbation, which indicates that our method performs better against stronger adversarial attacks. The results in the rest of the datasets for $\epsilon=4/255$ and $\epsilon=1/255$ perturbation are shown in the supplementary material.

\begin{table}

  \caption{The performance for $\epsilon=4/255$ under different shots and part datasets. The results of HEP are copied under different shots in
 the table for the convenience of comparison.}

  \label{tab:tab1}
\captionsetup{justification=centering}
\footnotesize
\centering
\resizebox{8.5cm}{!}{
\begin{tblr}{
  cell{1}{1} = {r=2}{l},
  cell{1}{2} = {r=2}{l},
  cell{1}{3} = {c=2}{},
  cell{1}{5} = {c=2}{},
  cell{1}{7} = {c=2}{},
  cell{1}{9} = {c=2}{},
  cell{3}{1} = {r=6}{},
  cell{9}{1} = {r=6}{},
  cell{15}{1} = {r=6}{},
  cell{21}{1} = {r=6}{},
  cell{1-38}{3-10} = {c},
  cell{27}{1} = {r=6}{},
  cell{33}{1} = {r=6}{},
  hline{1,39} = {-}{0.15em},
  hline{3,9,15,21,27,33}={-}{0.1em},
  hline{2} = {3-10}{0.03em},
  hline{8} = {2-10}{0.03em},
  hline{14,20,26,32,38} = {2-10}{0.03em},
}
Dataset        & Method  & 1-shot &      & 4-shots &      & 16-shots &      & All  &      \\
               &         & Acc.   & Rob.   & Acc.   & Rob.   & Acc.   & Rob.  & Acc.    & Rob.  \\
ImageNet       & HEP \cite{radford2021learning}     & 39.63  & 10.28  & \underline{39.63}  & 10.28  & 39.63  & 10.28  & 39.63  & 10.28  \\
               & PAFT \cite{chen2020adversarial}      & 9.23   & 3.29   & 15.93  & 5.37   & 31.92  & \underline{12.90}  & 17.12  & 7.26   \\
               & AVP \cite{chen2023visual}      & \underline{39.66}  & 10.41  & 39.58  & 10.97  & 39.61  & 11.18  & 39.53  & 11.32  \\
               & APT-CSC \cite{li2024one} & 18.56  & 4.08   & 29.88  & 6.99   & 35.10  & 8.46   & 39.91  & 11.87  \\
               & APT-UC \cite{li2024one}  & 37.88  & \underline{11.21}  & 39.52  & \underline{11.52}  & \underline{40.78}  & 12.17  & \underline{41.64}  & \underline{12.49}  \\
               & CAPT    & \textbf{46.81}  & \textbf{15.26}  & \textbf{55.90}  & \textbf{16.69}  & \textbf{60.98}  & \textbf{21.73}  & \textbf{65.67}  & \textbf{24.48}  \\
Caltech101     & HEP     & \underline{77.44}  & 42.76  & 77.44  & 42.76  & 77.44  & 42.76  & 77.44  & 42.76  \\
               & PAFT      & 51.32  & 32.04  & 72.55  & 46.36  & \underline{86.42}  & \underline{60.55}  & 88.85  & \underline{66.99}  \\
               & AVP      & 77.40  & \underline{43.12}  & 77.44  & 43.57  & 77.36  & 44.67  & 77.61  & 46.33  \\
               & APT-CSC & 61.81  & 33.24  & 78.67  & 46.79  & \underline{86.42}  & 57.30  & \underline{89.63}  & 65.59  \\
               & APT-UC  & 75.90  & 42.66  & \underline{82.46}  & \underline{50.72}  & 86.37  & 56.52  & 88.75  & 63.51  \\
               & CAPT    & \textbf{79.92}  & \textbf{57.52}  & \textbf{86.86}  & \textbf{64.91}  & \textbf{92.54}  & \textbf{73.43}  & \textbf{95.01}  & \textbf{79.43}  \\
OxfordPets     & HEP     & \underline{61.49}  & 14.34  & \underline{61.49}  & 14.34  & 61.49  & 14.34  & 61.49  & 14.34  \\
               & PAFT      & 12.33  & 3.07   & 36.16  & 10.17  & 63.47  & 19.45  & 71.46  & \underline{25.60}  \\
               & AVP      & 60.89  & \underline{14.83}  & 61.02  & \underline{14.94}  & 61.00  & 15.24  & 60.89  & 16.63  \\
               & APT-CSC & 35.87  & 6.75   & 51.73  & 9.90   & 66.07  & 17.19  & 72.36  & 24.52  \\
               & APT-UC  & 57.49  & 14.31  & 61.21  & 14.71  & \underline{67.77}  & \underline{20.29}  & \underline{72.43}  & 24.72  \\
               & CAPT    & \textbf{63.40}   & \textbf{17.88}  & \textbf{67.83}  & \textbf{27.26}  & \textbf{79.99}  & \textbf{40.94}  & \textbf{87.11}  & \textbf{49.50}   \\
StanfordCars   & HEP     & 10.33  & 0.92   & 10.33  & 0.92   & 10.33  & 0.92   & 10.33  & 0.92   \\
               & PAFT      & 5.98   & \underline{2.02}   & 14.09  & 3.84   & 34.27  & \underline{11.09}  & 41.03  & \underline{13.75}  \\
               & AVP      & 10.38  & 0.95   & 10.65  & 1.08   & 11.51  & 1.52   & 11.70  & 1.72   \\
               & APT-CSC & 12.48  & 1.73   & \underline{26.23}  & 4.68   & \underline{42.29}  & 9.85   & \underline{48.83}  & 12.83  \\
               & APT-UC  & \underline{12.54}  & 1.94   & 25.72  & \underline{4.95}   & 32.98  & 7.82   & 37.58  & 9.05   \\
               & CAPT    & \textbf{36.93}  & \textbf{8.75}   & \textbf{56.59}  & \textbf{16.86}  & \textbf{74.59}  & \textbf{26.17}  & \textbf{81.17}  & \textbf{31.60}   \\
Food101        & HEP     & \underline{21.70}  & 3.19   & 21.70  & 3.19   & 21.70  & 3.19   & 21.70  & 3.19   \\
               & PAFT      & 5.98   & 1.16   & 14.13  & 3.08   & 30.45  & \underline{8.23}   & 36.87  & 14.60  \\
               & AVP      & 20.28  & 3.12   & 20.57  & 3.27   & 20.69  & 3.50   & 24.50  & 5.98   \\
               & APT-CSC & 11.41  & 1.25   & 20.75  & 2.65   & \underline{33.15}  & 6.72   & \underline{42.75}  & \underline{14.71}  \\
               & APT-UC  & 19.85  & \underline{3.65}   & \underline{23.51}  & \underline{4.05}   & 30.56  & 7.94   & 35.52  & 12.98  \\
               & CAPT    & \textbf{30.45}  & \textbf{9.09}   & \textbf{42.81}  & \textbf{13.45}  & \textbf{62.01}  & \textbf{21.97}  & \textbf{80.91}  & \textbf{36.09}  \\
FGVCAircraft   & HEP     & 7.02   & 0.48   & 7.02   & 0.48   & 7.02   & 0.48   & 7.02   & 0.48   \\
               & PAFT      & 6.38   & \underline{1.96}   & 12.70  & 2.96   & 26.35  & \underline{7.94}   & 29.28  & \underline{10.08}  \\
               & AVP      & 6.42   & 0.42   & 6.78   & 0.60   & 7.71   & 1.05   & 8.01   & 1.32   \\
               & APT-CSC & \underline{9.56}   & 0.94   & \underline{16.86}  & 2.50   & \underline{28.64}  & 6.82   & \underline{31.56}  & 8.91   \\
               & APT-UC  & 2.81   & 0.75   & 11.21  & \underline{3.27}   & 17.68  & 5.73   & 20.89  & 7.02   \\
               & CAPT    & \textbf{12.99}  & \textbf{4.23}   & \textbf{21.27}  & \textbf{4.62}   & \textbf{40.35}  & \textbf{19.14}  & \textbf{50.47}  & \textbf{25.05} 
\end{tblr}
}
\end{table}

\begin{table}
  \caption{The performance for $\epsilon=1/255$ under different shots and part datasets.}
  \centering
  \label{tab:tab2}
  \captionsetup{justification=centering}
  \footnotesize
\resizebox{8cm}{!}{
\centering
\begin{tblr}{
  cell{1}{1} = {r=2}{},
  cell{1}{2} = {r=2}{},
  cell{1}{3} = {c=2}{},
  cell{1}{5} = {c=2}{},
  cell{1}{7} = {c=2}{},
  cell{1}{9} = {c=2}{},
  cell{3}{1} = {r=6}{},
  cell{9}{1} = {r=6}{},
  cell{15}{1} = {r=6}{},
  cell{21}{1} = {r=6}{},
  cell{1-38}{3-10} = {c},
  cell{27}{1} = {r=6}{},
  cell{33}{1} = {r=6}{},
  hline{1,39} = {-}{0.15em},
  hline{3,9,15,21,27,33}={-}{0.1em},
  hline{2} = {3-10}{0.03em},
  hline{8} = {2-10}{0.03em},
  hline{14,20,26,32,38} = {2-10}{0.03em},
}
Dataset        & Method  & 1-shot &      & 4-shots &      & 16-shots &      & All  &      \\
               &         & Acc.   & Rob.   & Acc.   & Rob.   & Acc.   & Rob.  & Acc.    & Rob.   \\
ImageNet       & HEP     & \underline{55.16}  & \underline{38.49}  & 55.16  & 38.49  & 55.16  & 38.49  & 55.16  & 38.49  \\
               & PAFT      & 19.75  & 14.03  & 32.74  & 22.87  & 52.42  & \underline{39.12}  & 38.93  & 29.36  \\
               & AVP      & 55.15  & \textbf{38.66}  & 55.17  & \underline{38.78}  & 55.25  & 38.77  & 55.29  & 38.85  \\
               & APT-CSC & 31.92  & 20.48  & 45.51  & 29.75  & 50.84  & 33.21  & 57.65  & 40.29  \\
               & APT-UC  & 54.55  & 37.93  & \underline{56.42}  & \textbf{39.79}  & \underline{58.02}  & \textbf{40.83}  & \underline{58.56}  & \underline{41.53}  \\
               & CAPT    & \textbf{55.39}  & 29.87  & \textbf{60.66}  & 34.24  & \textbf{64.34}  & 38.93  & \textbf{68.77}  & \textbf{43.07}  \\
Caltech101     & HEP     & 83.94  & \textbf{74.00}  & 83.94  & 74.00  & 83.94  & 74.00  & 83.94  & 74.00  \\
               & PAFT      & 69.86  & 61.14  & 83.16  & 72.98  & 92.49  & \underline{85.35}  & \underline{93.87}  & \textbf{88.15}  \\
               & AVP      & 83.81  & \underline{74.00}  & 83.85  & 74.40  & 83.81  & 74.73  & 83.85  & 75.01  \\
               & APT-CSC & 75.21  & 63.77  & 86.21  & 75.42  & 91.24  & 84.06  & 93.47  & 86.77  \\
               & APT-UC  & \underline{86.04}  & 73.06  & \underline{90.30}  & \textbf{81.01}  & \underline{92.66}  & \textbf{85.40}  & 93.67  & 87.14  \\
               & CAPT    & \textbf{88.48}  & 73.83  & \textbf{90.34}  & \underline{79.15}  & \textbf{94.12}  & 84.34  & \textbf{95.50}  & \underline{87.91}  \\
OxfordPets     & HEP     & \underline{74.87}  & \underline{58.63}  & 74.87  & \underline{58.63}  & 74.87  & 58.63  & 74.87  & 58.63  \\
               & PAFT      & 29.30  & 19.43  & 58.11  & 42.06  & 80.40  & \underline{63.21}  & 86.05  & \textbf{69.69}  \\
               & AVP      & 74.73  & 58.11  & 74.65  & 58.08  & 74.76  & 58.27  & 74.98  & 58.52  \\
               & APT-CSC & 60.81  & 42.85  & 71.79  & 51.35  & 80.21  & 60.67  & 86.29  & 69.47  \\
               & APT-UC  & \textbf{79.67}  & \textbf{61.41}  & \textbf{81.58}  & \textbf{62.44}  & \underline{83.05}  & \textbf{65.55}  & \underline{86.45}  & \underline{69.53}  \\
               & CAPT    & 74.13  & 41.53  & \underline{78.58}  & 53.99  & \textbf{84.85}  & 61.16  & \textbf{88.36}  & 69.15  \\
StanfordCars   & HEP     & 25.31  & 12.30  & 25.31  & 12.30  & 25.31  & 12.30  & 25.31  & 12.30  \\
               & PAFT      & 15.63  & 9.82   & 33.63  & 21.70  & 62.12  & \underline{43.63}  & 69.12  & \underline{50.68}  \\
               & AVP      & 25.52  & 12.49  & 27.92  & 15.09  & 30.99  & 18.67  & 31.84  & 19.25  \\
               & APT-CSC & 24.79  & 13.06  & 44.17  & 26.23  & \underline{64.82}  & 43.40  & \underline{72.18}  & 50.01  \\
               & APT-UC  & \underline{36.35}  & \underline{19.15}  & \underline{48.76}  & \underline{26.89}  & 60.23  & 36.87  & 63.71  & 38.96  \\
               & CAPT    & \textbf{46.04}  & \textbf{22.94}  & \textbf{60.72}  & \textbf{35.51}  & \textbf{77.02}  & \textbf{51.37}  & \textbf{82.76}  & \textbf{56.83}  \\
Food101        & HEP     & \underline{44.99}  & \textbf{26.25}  & 44.99  & 26.25  & 44.99  & 26.25  & 44.99  & 26.25  \\
               & PAFT      & 14.06  & 8.55   & 31.27  & 17.82  & 51.82  & 32.50  & 64.33  & 45.72  \\
               & AVP      & 43.16  & 25.39  & 43.44  & \underline{26.34}  & 43.98  & 27.34  & 48.14  & 32.24  \\
               & APT-CSC & 23.92  & 12.49  & 36.35  & 18.63  & 52.03  & 30.57  & \underline{66.56}  & \underline{46.24}  \\
               & APT-UC  & \textbf{45.67}  & \underline{26.05}  & \underline{45.75}  & 25.28  & \underline{55.66}  & \underline{33.73}  & 63.56  & 42.93  \\
               & CAPT    & 43.11  & 22.04  & \textbf{54.38}  & \textbf{28.79}  & \textbf{66.94}  & \textbf{38.52}  & \textbf{82.26}  & \textbf{54.77}  \\
FGVCAircraft   & HEP     & 12.48  & 5.88   & 12.48  & 5.88   & 12.48  & 5.88   & 12.48  & 5.88   \\
               & PAFT      & 12.75  & \underline{8.37}   & 20.01  & \underline{12.15}  & 35.43  & \underline{21.90}  & 39.51  & 25.92  \\
               & AVP      & 12.33  & 6.00   & 12.93  & 7.14   & 14.61  & 9.48   & 15.12  & 9.93   \\
               & APT-CSC & \underline{14.61}  & 7.38   & \underline{23.34}  & \textbf{12.30}  & \underline{36.81}  & 21.24  & \underline{41.73}  & \underline{26.22}  \\
               & APT-UC  & 13.89  & 7.71   & 21.21  & 10.92  & 28.47  & 15.39  & 31.95  & 18.84   \\
               & CAPT    & \textbf{14.61}  & \textbf{10.02}  & \textbf{23.67}  & 10.74  & \textbf{42.57}  & \textbf{27.18}  & \textbf{52.09}  & \textbf{35.91} 
\end{tblr}
}
\end{table}

\subsection{Cross Dataset Generalization}
This section assesses the generalization capacity of our model on distribution shifts and cross-dataset scenarios. We use ImageNet as the source dataset for adversarial prompt tuning. Then, we evaluate the performance of these ImageNet-adapted models on the target datasets with the same classes yet different data distributions and the target datasets with different classes. Specifically, follow Li et al. \cite{li2024one}. , we use three ImageNet shift datasets, ImageNet-V2 \cite{recht2019imagenet}, ImageNet-Sketch and ImageNet-R \cite{hendrycks2021many} to represent different kinds of distribution shift. We use the rest ten datasets for the cross-dataset test.

\begin{table}
  \renewcommand\arraystretch{1.2}
  \caption{The generalization of the prompts learned by our method on ImageNet to datasets with input distribution shifts. The results for both variants of our method are reported for the checkpoints trained with 16 shots and $\epsilon=4/255$.}
  \label{tab:freq}
  \resizebox{8.5cm}{!}{
  \begin{tabular}{l|cc|cccccc}
    \Xhline{0.6pt} 
        \multirow{3}{*}{Method} & \multicolumn{2}{c|}{Source} & \multicolumn{6}{c}{Distribution Shifts} \\ 
        ~ & \multicolumn{2}{c|}{ImageNet} & \multicolumn{2}{c}{ImageNet-V2} & \multicolumn{2}{c}{ImageNet-Sketch} & \multicolumn{2}{c}{ImageNet-R} \\ \cline{2-9}
        ~ & Acc. & Rob. & Acc. & Rob. & Acc. & Rob. & Acc. & Rob. \\ 
    \hline
        HEP \cite{radford2021learning} & 39.86 & 10.28 & 32.74 & 7.49 & 17.40 & 7.21 & 21.46 & 5.80 \\ 
        AVP \cite{chen2023visual} & 39.61 & 11.18 & 32.68 & 8.12 & 17.39 & 7.69 & 21.47 & 6.25 \\ 
        PAFT \cite{chen2020adversarial} & 31.92 & 12.90 & 25.55 & 9.52 & 10.02 & 5.05 & 13.34 & 4.55 \\ 
        APT-CSC \cite{li2024one} & 37.18 & 9.49 & 28.93 & 6.65 & 12.72 & 4.83 & 15.06 & 3.57 \\ 
        APT-UC \cite{li2024one} & 40.80 & 12.33 & 33.20 & 9.04 & 18.35 & 8.04 & 22.66 & 6.97 \\ 
    \hline
        Ours w/o CG & 60.53 & 18.45 & 51.04 & 14.08 & \textbf{27.26} & 15.24 & 31.08 & 12.36 \\ 
        Ours with CG & \textbf{60.98} & \textbf{21.73} & \textbf{51.05} & \textbf{17.04} & 27.01 & \textbf{15.67} & \textbf{31.45} & \textbf{13.38} \\
  \Xhline{0.6pt} 
\end{tabular}
}
\end{table}

We undertake studies on our technique with and without the adaptive consistency led by the pre-trained frozen CLIP, in addition to doing research on alternative baseline methods. Based on the results presented in Tab \ref{tab:freq}, it is evident that our methodology surpasses all other approaches by a large margin, even when the consistency guidance from the pre-trained frozen CLIP is not included. This is true for both zero-shot accuracy and robustness in the distribution shifts scenario. With the support of the generalized information from the pre-trained frozen CLIP, the robust generalization of the model was further enhanced without compromising the accuracy of clean instances. This was accomplished without compromising efficiency. When taken as a whole, the result demonstrates that our approach is effective in improving the robust generalization of the model that has been partially trained. It is mentioned in the supplemental material that the findings of the cross-dataset test were obtained.

\subsection{Ablation Study}
To reveal the effect of different terms in our objective function, we conduct experiments with different objective functions for adversarial prompt tuning. We set 5 different cases including $\rm CE_{Adv}$,$\rm CE_{clean}$,
${\rm CE_{clean}} + L_{\rm cons-train}$,${\rm CE_{clean}} + L_{\rm cons-frz}$ and All. $\rm CE_{Adv}$ denotes the cross-entropy of the prediction of the adversarial examples with the label, which is the same loss function as APT \cite{li2024one}.  $\rm CE_{clean}$ denotes the cross-entropy of the prediction of the clean examples with the label without adopting adversarial training. $L_{\rm cons-train}$ and $L_{\rm cons-frz}$ denote consistency loss between clean and adversarial inputs of fine-tuned model and frozen pre-trained model respectively as mentioned above. All the experiments in this section use 16 shots on the StanfordCars dataset. According to the result in Tab \ref{tab:tab3}, we could see that solely using $\rm CE_{Adv}$ achieves better robustness. However, its accuracy on clean examples drops significantly as it disregards the clean ones. Both methods combining $\rm CE_{clean}$ with $L_{\rm cons-train}$ or $L_{\rm cons-frz}$ improve the robustness of the model to a great extent while ensuring the accuracy compared to using $\rm CE_{clean}$ independently. Combining $\rm CE_{clean}$ with $L_{\rm cons-train}$ and $L_{\rm cons-frz}$ together achieves a greater improvement in robustness and even a slightly higher accuracy. The results of the ablation study once again demonstrate that our approach achieves a balance in improving the robust generalization of the model while retaining the accuracy of clean examples.
 
\section{Conclusion}
In this work, we focus on the problem of over-fitting that occurs in adversarial prompt tuning. This problem limits the model's capacity to generalize on both clean and adversarial scenarios. We present the CAPT framework in order to address this issue. This framework makes use of multi-modal prompt learning in order to enhance the alignment of text-image features for adversarial cases and to raise consistency across both clean and adversarial inputs. In addition, we enhanced the robust generalization on adversarial scenarios by using a pre-trained frozen CLIP, which ensured that the correctness of clean examples was preserved. 
As a consequence of our rigorous testing, we have determined that CAPT is superior in terms of both its performance inside the distribution and its capacity to generalize both in instances when there is a shift in the distribution and across datasets. In this paper, we take a novel and efficient approach to addressing the problem of adversarial rapid tuning, which is characterized by over-fitting.

\begin{table}
  \caption{Ablation study of different objective function} 
  \label{tab:tab3} 
    \begin{tabular}{*{4}{c}|*{2}{c}}
        \Xhline{0.6pt} 
        \multicolumn{4}{c|}{Method} & \multicolumn{2}{c}{Result} \\
        $\rm CE_{Adv}$ & $\rm CE_{clean}$ & $\rm L_{cons-train}$ & $\rm L_{cons-frz}$ & Acc. & Rob.    \\
        \hline
        \checkmark &              &                  &                & 44.35     & \textbf{36.70}   \\
        
                   & \checkmark   &                  &                & 74.97     & 8.47   \\
        
                   & \checkmark   & \checkmark       &                & 73.88     & 33.24   \\
        
                   & \checkmark   &                  & \checkmark     & 75.69     & 29.44   \\
        
                   & \checkmark   & \checkmark       & \checkmark     & \textbf{76.07}     & 35.65   \\
        \Xhline{0.6pt}
    \end{tabular}
\end{table}


\newpage
\bibliographystyle{ACM-Reference-Format}
\bibliography{sample-base}










\end{document}


\title{Supplementary Materials: Revisiting the Robust Generalization of Adversarially Prompt Tuning}


\author{Fan Yang, Mingxuan Xia, Sangzhou Xia, Chicheng Ma, Hui Hui}

\maketitle

\section{Related works}
\subsection{Adversarial Attack on CLIP}
A common strategy to generate adversarial examples for Vision Language Models (VLMs) involves searching for a perturbation $\boldsymbol{\delta_i}$ that maximizes the dissimilarity, typically the cosine dissimilarity, between the image feature $\boldsymbol{z_v^i}$ and the text feature of the corresponding ground-truth class prompt $\boldsymbol{z_t^{y_j}}$. This perturbation $\boldsymbol{\delta_i}$ is subject to a constraint that it lies within an $\boldsymbol{\epsilon}$-ball defined by the $p$-norm. With the learnable prompt $\boldsymbol{P}$ in adversarial prompt tuning(APT), the objective can be expressed as:

\begin{equation}
\mathop{\arg\max}_{\|\boldsymbol{\delta_i}\|_p \leq \boldsymbol{\epsilon}} \mathcal{L}(x_i + \boldsymbol{\delta_i}, \boldsymbol{t}, y_j; \boldsymbol{\theta_\mathcal{I}}, \boldsymbol{\theta_\mathcal{T}}, \boldsymbol{P})
\end{equation}

This formulation diverges from traditional approaches due to the inclusion of the text encoder $\boldsymbol{\theta_\mathcal{T}}$. During the generation of the adversarial examples, all the parameters in VLMs are fixed including the multi-modal prompts. Due to the existence of text prompts in the text encoder, the text feature of the corresponding ground-truth class changes during the adversarial prompt tuning. The adversarial attack described above is delineated in Algorithm \ref{tab: Alg1}.

\subsection{Adversarial Visual Prompting}
AVP \cite{chen2023visual} combines visual prompting \cite{bahng2022exploring} with adversarial
training to counteract adversarial perturbations. The visual
prompt perturbation, parameterized by $\phi$ , is applied to the
input image, $x$, so that the prompted image is given by
$x_{vp} = x + \phi$. AVP optimizes visual prompt $\phi$ to jointly
minimize both clean and adversarial losses:
$$arg \min _{\phi} \mathcal{L}\left(x+\phi, t, y ; \theta_{v}, \theta_{t}\right)+\mathcal{L}\left(x+\delta+\phi, t, y ; \theta_{v}, \theta_{t}\right)$$
where $\delta$ is generated by Algorithm  \ref{tab: Alg1} independent of $\phi$.

\subsection{Partial Adversarial Fine-Tuning}
PAFT \cite{chen2020adversarial} discards the text encoder branch of CLIP and
attaches an extra linear layer, parameterized by $\theta_{l}$
, on top
of the frozen image encoder, $\theta_{v}$, to form a new classifier.
The output of the linear classifier has the same dimension
as the number of classes. PAFT optimizes $\theta_{l}$
to minimize
the adversarial loss:
$$arg \min _{\theta_{l}} \mathcal{L}\left({x}+\theta_{v}, y ; \theta_{v}, \theta_{l}\right)$$
where $\delta$ is generated using the conventional PGD attack on
the new classifier.


\section{Experiment Setting}
\subsection{Data Setting} 
\textbf{Data.} The selection of the 11 datasets was designed to establish a comprehensive benchmark that spans a wide range of vision-related tasks. This includes generic object classification, scene recognition, action categorization, fine-grained classification, texture recognition, and the analysis of satellite images. Consistent with the approach taken by Zhou et al. \cite{zhou2022learning}, these datasets were divided into training and testing sets. To maintain uniformity across evaluations, the N-shot images were sampled once and then fixed, ensuring that all methods under comparison are trained on identical datasets, thus guaranteeing a fair assessment.

\textbf{Model.} The text encoder is the default pre-trained model
from CLIP \cite{radford2021learning}. For both image and text branches, we
adopt the same data pre-processing as CLIP \cite{radford2021learning}.

\begin{table}
  \caption{Datasets statistics and their prompts}
  \label{tab: HEP}
  \resizebox{8cm}{!}{
  \begin{tabular}{ccl}
    \toprule
    Dataset & Classes & Hand-Engineered Prompt\\
    \midrule
    ImageNet & 1000 &  a photo of a [CLASS] \\
    OxfordPets & 37 & a photo of a [CLASS], a type of pet \\
    StanfordCars &  196 &  a photo of a [CLASS] \\
    Flowers102 & 102 &  a photo of a [CLASS], a type of flower \\
    Food101 & 101 & a photo of a [CLASS], a type of food \\
    FGVCAircraft &  100 & a photo of a [CLASS], a type of aircraft \\
    SUN397 & 397 & a photo of a [CLASS] \\
    DTD & 47 & [CLASS] texture \\
    EuroSAT & 10 & a centered satellite photo of a [CLASS] \\
    UCF101 & 101 & a photo of a person doing [CLASS] \\
    ImageNetV2 & 1000 & a photo of a [CLASS] \\
    ImageNet-Sketch & 1000 & a photo of a [CLASS] \\
    ImageNet-A & 200 & a photo of a [CLASS] \\
    ImageNet-R & 200 & a photo of a [CLASS] \\
  \bottomrule
\end{tabular}
}
\end{table}

\subsection{Baseline Setting} 
\textbf{HEP} Hand-Engineered Prompts (HEP) which was originally proposed in CLIP and has been widely used in other works \cite{zhao2024evaluating,zhou2022learning,zhou2022conditional}. The specific prompts used for each dataset are described in Tab. \ref{tab: HEP}.

\textbf{APT} was trained by Stochastic Gradient Descent (SGD) using a cosine learning rate schedule with an initial learning
rate of 0.002. The batch size was 32. For ImageNet, the
number of epochs was 20, 20, 50 and 20 for 1, 4, 16 and
100 shots respectively, and for other datasets was 50, 100,
200, 200 for 1, 4, 16 and all shots respectively. The number
of epochs used with ImageNet was reduced due to limited
computational resources. Results were reported for the last
checkpoint.

\textbf{AVP}  was implemented in the mode of padding with a
prompt size of 52 to match the number of parameters of our
method for the unified context. Although a class-specific
variant of AVP was proposed in \cite{chen2023visual} and was observed to
outperform its unified variant on CIFAR10 with ResNet18,
we found that the class-specific variant of AVP generalized poorly to our experimental set-ups, i.e., CLIP plus
11 diverse datasets. We therefore decided to use the unified variant of AVP. Following \cite{chen2023visual}, a cosine learning rate
schedule with an initial learning rate of 0.1 was used. The
number of epochs, corresponding to 1/4/16/all shots, was
20/50/100/100 for non-ImageNet datasets and 20/20/50/20
for ImageNet. The implementation was based on the open-source code of Chen et al. \cite{chen2023visual}.

\textbf{PAFT} was trained, following Chen et al. \cite{chen2020adversarial}, for
20/50/100/100 epochs with an initial learning rate of 0.1
decayed by 0.1 at epochs 5/15/30/30 and 10/25/50/50 for
1/4/16/all shots, respectively, for non-ImageNet datasets.
For ImageNet, the number of training epochs was
20/20/50/20 for 1/4/16/100 shots respectively.

\section{Method description}
For a better understanding of our proposed adaptive Consistency-guided Adversarial Prompt Tuning (CAPT), we describe the pipeline of our approach in Algorithm \ref{tab: Alg2}.

\section{Additional Results}
\subsection{In-Distribution Performance}
\begin{figure*}[!t]
\centerline{\includegraphics[width=0.85\textwidth]{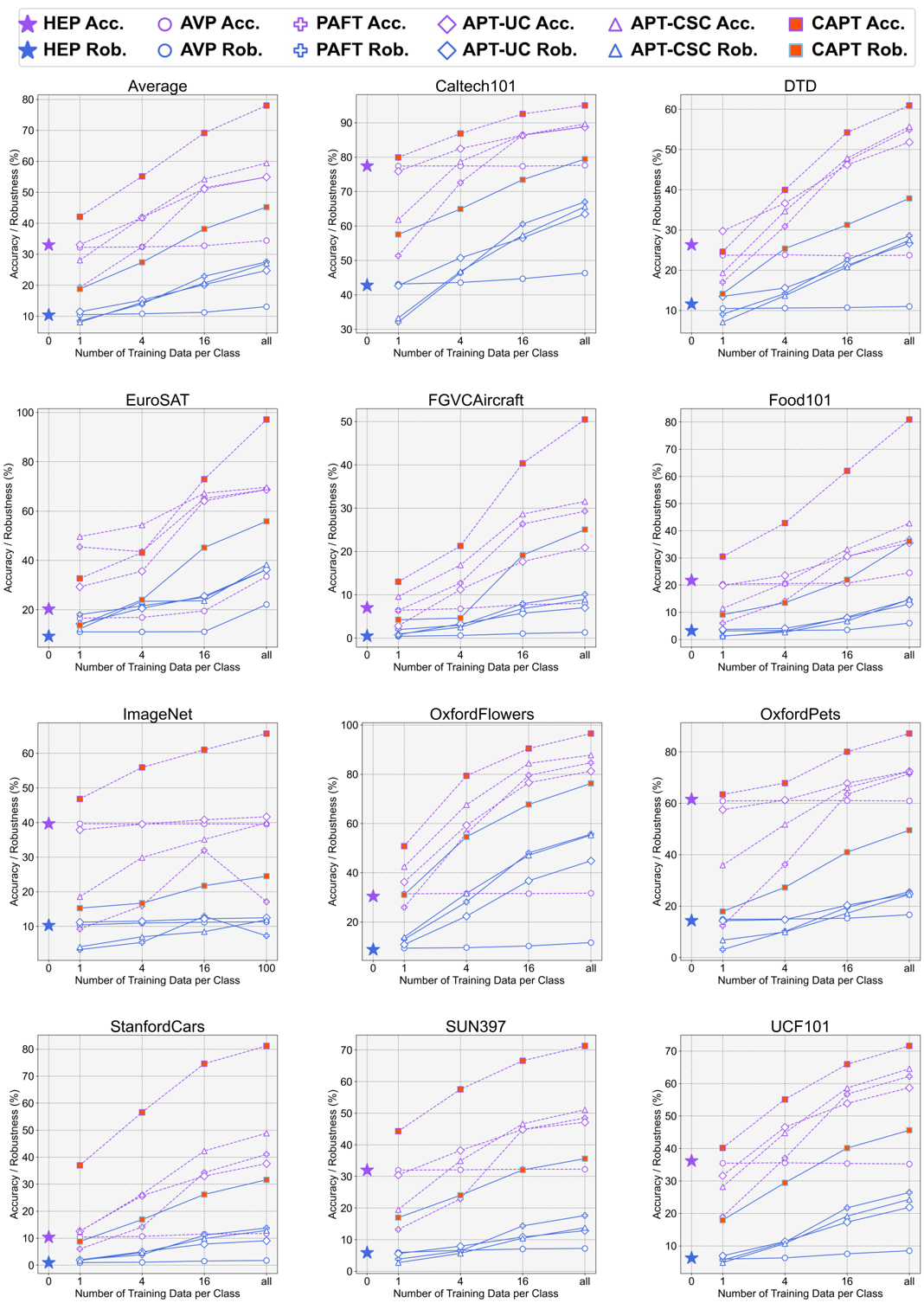}}
\caption{The in-distribution performance on 11 datasets and the averaged performance under different shots. $\epsilon=4/255$}

\label{eps4}
\end{figure*}

\begin{figure*}[!t]
\centerline{\includegraphics[width=0.85\textwidth]{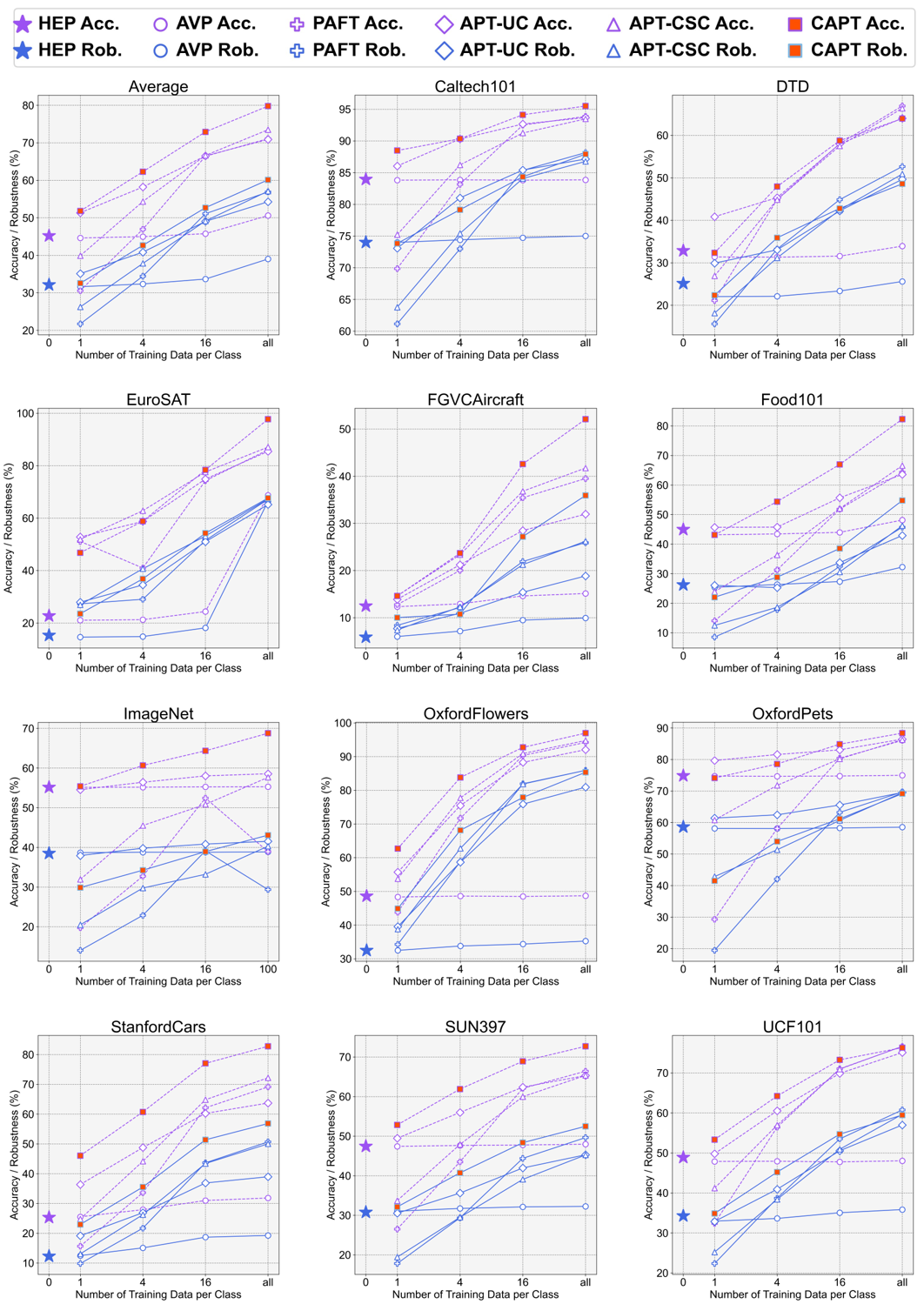}}
\caption{The in-distribution performance on 11 datasets and the averaged performance under different shots. $\epsilon=1/255$}
\label{eps1}
\end{figure*}

Fig. \ref{eps4} and Fig. \ref{eps1}shows the results for $\epsilon=4/255$ and $\epsilon=1/255$ respectively. From the figure, we can see that most of the accuracy and robustness of our proposed method are significantly better than those of other methods under different shots of different datasets, especially for $\epsilon=4/255$. Besides, a specific comparison of different text prompting methods on the performance averaged over 11 datasets for various perturbation budgets $\epsilon$, and shots are shown in Tab. \ref{tab:2}. On average, our method also significantly outperforms other methods.

\subsection{Zero-shot Performance}
In this section, we evaluate the zero-shot capabilities following the evaluation protocol of  TeCoA \cite{mao2022understanding}. Our approach involved tuning the models using ImageNet (the source dataset) before conducting assessments on the target datasets, which consist of different classes, specifically the remaining ten out of the original eleven datasets. It's important to note that PAFT \cite{chen2020adversarial}is unable to process new classes not encountered during its training phase due to its inflexible, hard-coded linear layer. Consequently, it is unsuitable for this type of evaluation. This limitation is also applied to the APT-CSC\cite{li2024one}. 

In Tab. \ref{tab:zero-shot}, CAPT with Consistency Guided achieves the highest robustness on all datasets except EuroSAT. We believe that the reason our method fails to generalize on EuroSAT is due to the significant difference in data distribution between it and the source dataset, ImageNet. Nevertheless, our approach still achieved significant improvements on zero-shot accuracy and robustness for the other nine datasets, which demonstrates the superiority on improving the accuracy and robustness generalization of the pre-trained model.

\setcounter{table}{0}
\begin{table*}[h]
\begin{flushleft} 
\Large 
\caption{Adversarial attack on CLIP}
\label{tab: Alg1}
\begin{tabularx}{\textwidth}{Xl}
\toprule
\multicolumn{2}{l}{\textbf{Algorithm 1} Pseudo-code for adversarial attack on CLIP.} \\
\midrule
\quad \textbf{Input:} input images $\boldsymbol{x}$, label $\boldsymbol{y}$, text embedding  $\boldsymbol{t}$, pre-trained CLIP parameters $\boldsymbol{\theta}=\{\boldsymbol{\theta}_\mathcal{I},\boldsymbol{\theta}_\mathcal{T}\}$, learnable prompts \\
\quad $\boldsymbol{P}$, perturbation bound $\epsilon$, step size for perturbing image $\alpha$ and step number $K$. \\
\quad \textbf{Output:} adversarial examples $\boldsymbol{x_a}$ \\
\quad \textbf{function} ATTACK$(\boldsymbol{x}, \boldsymbol{y}, \boldsymbol{t}, \boldsymbol{\theta}, \boldsymbol{P}, \alpha, \epsilon, K)$ \\
\quad \quad $\boldsymbol{\delta} \gets \text{uniform}(-\epsilon, \epsilon)$  \\
\quad \quad \textbf{for} $1 \rightarrow K$ \textbf{do} \\
\quad \quad \quad $\boldsymbol{x'} \gets \min(0, \max(\boldsymbol{x} + \boldsymbol{\delta}, 1))$ \\
\quad \quad \quad $L \gets \mathcal{L}(x_i + \boldsymbol{\delta_i}, \boldsymbol{t}, \boldsymbol{y}; \boldsymbol{\theta}_\mathcal{I}, \boldsymbol{\theta}_\mathcal{T}, \boldsymbol{P})$ \\
\quad \quad \quad $\boldsymbol{\delta} \gets \min(-\epsilon, \max(\boldsymbol{\delta} + \alpha \cdot\text{SIGN}(\nabla_{\boldsymbol{x}}L), \epsilon))$ \\
\quad \quad \quad \textbf{end if} \\
\quad \quad \textbf{end for} \\
\quad \quad \textbf{return} $\min(0, \max(\boldsymbol{x} + \boldsymbol{\delta}, 1))$ \\
\quad \textbf{end function} \\ 
\bottomrule 
\end{tabularx}
\end{flushleft}
\end{table*}

\vspace{2cm}

\begin{table*}[h]
\begin{flushleft} 
\Large 
\caption{Pipeline of adaptive Consistency-guided Adversarial Prompt Tuning}
\label{tab: Alg2}
\begin{tabularx}{\textwidth}{Xl}
\toprule
\multicolumn{2}{l}{\textbf{Algorithm 2} adaptive Consistency-guided
 Adversarial Prompt Tuning (CAPT).} \\
\midrule

\quad \textbf{Input:} input images $\boldsymbol{x}$, label $\boldsymbol{y}$, text embedding  $\boldsymbol{t}$, pre-trained CLIP parameters $\boldsymbol{\theta}=\{\boldsymbol{\theta}_\mathcal{I},\boldsymbol{\theta}_\mathcal{T}\}$, frozen CLIP \\ 
\quad  parameters $\boldsymbol{t}$, pre-trained CLIP parameters $\boldsymbol{\theta}_{\rm frz}= $ 
$\{
\boldsymbol{\theta}_{\mathcal{I}_{\rm frz}}, \boldsymbol{\theta}_{\mathcal{T}_{\rm frz}}
\}$
learnable multi-modal prompts 
$\boldsymbol{P}=\{\boldsymbol{P}_\mathcal{I},\boldsymbol{P}_\mathcal{T}\}$, \\
\quad perturbation bound $\epsilon$, step size for perturbing image $\alpha$, step number $K$, learning rate $\eta$ and weight parameter $\lambda$. \\
\quad \textbf{for} $1 \rightarrow \textbf{EPOCH}$ \textbf{do} \\
\quad \quad \textbf{for} $\boldsymbol{x},\boldsymbol{y}$ in mini-batch \textbf{do} \\
\quad \quad \quad\textcolor{green}{\# Generate adversarial examples using Algorithm \ref{tab: Alg1}}  \\
\quad \quad \quad $\boldsymbol{x_a}=$ATTACK$(\boldsymbol{x}, \boldsymbol{y}, \boldsymbol{t}, \boldsymbol{\theta}, \boldsymbol{P}, \alpha, \epsilon, K)$ \\
\quad \quad \quad\textcolor{green}{\# Extract multi-modal features from prompt tuning model}  \\
\quad \quad \quad $\boldsymbol{z}^I=f_{\mathcal{I}}(\boldsymbol{x},\boldsymbol{\theta}_{\mathcal{I}},\boldsymbol{P}_{\mathcal{I}})$,\quad
$\boldsymbol{z}^I_a=f_{\mathcal{I}}(\boldsymbol{x_a},\boldsymbol{\theta}_{\mathcal{I}},\boldsymbol{P}_{\mathcal{I}})$,\quad
$\boldsymbol{z}^T=f_{\mathcal{T}}(\boldsymbol{x},\boldsymbol{\theta}_{\mathcal{T}},\boldsymbol{P}_{\mathcal{T}})$ \\
\quad \quad \quad\textcolor{green}{\# Extract multi-modal features from frozen model}  \\

\quad  \quad \quad $\boldsymbol{z}_{\rm frz}^I=f_{\mathcal{I}_{\rm frz}}(\boldsymbol{x},\boldsymbol{\theta}_{\mathcal{I}_{\rm frz}})$,\quad
$\boldsymbol{z}^T=f_{\mathcal{T}_{\rm frz}}(\boldsymbol{x},\boldsymbol{\theta}_{\mathcal{T}_{\rm frz}})$ \\

\quad \quad \quad\textcolor{green}{\# Compute cross-entropy loss on clean examples}  \\
\quad \quad \quad$L_{\rm CE}={\rm CE}({\rm sft}({\boldsymbol{z}}^I \cdot \boldsymbol{z}^T / \tau), \boldsymbol{y})$ \\ 

\quad \quad \quad\textcolor{green}{\# Compute adaptive consistency-guided loss }  \\

\quad \quad \quad $L_{\rm cons-train} = {\rm KL}(({\rm sft} (\boldsymbol{z}^{I_a} \cdot \boldsymbol{z}^T / \tau),{\rm sft}(\boldsymbol{z}^I \cdot \boldsymbol{z}^T / \tau)))$ \\

\quad \quad \quad $L_{\rm cons-frz} = {\rm KL}(({\rm sft} (\boldsymbol{z}^{I_a} \cdot \boldsymbol{z}^T / \tau),{\rm sft}(\boldsymbol{z}_{\rm frz}^I \cdot \boldsymbol{z}_{\rm frz}^T / \tau))$ \\

\quad \quad \quad    $\alpha_{\rm cons}  = {{\rm exp}({\rm CE}({\rm sft}({\boldsymbol{z}}_{\rm frz}^I \cdot {\boldsymbol{z}}_{\rm frz}^T / \tau), \boldsymbol{y}))} / {[{\rm exp}({\rm CE}({\rm sft}({\boldsymbol{z}}_{\rm frz}^I \cdot \boldsymbol{z}_{\rm frz}^T / \tau), \boldsymbol{y}) + {\rm exp}({\rm CE}({\rm sft}({\boldsymbol{z}}^I \cdot \boldsymbol{z}^T / \tau), \boldsymbol{y})]}$ \\

\quad \quad \quad    $L_{\rm adv-cons}  =  (1-\alpha_{\rm cons})L_{\rm cons-train} + \alpha_{\rm cons} L_{\rm cons-frz}$ \\

\quad \quad \quad\textcolor{green}{\# Compute the final loss and update learnable multi-modal prompt}  \\
\quad \quad \quad $L_{\rm CAPT}=L_{\rm CE} + \lambda L_{\rm adv-cons}$ \\
\quad \quad \quad $ \boldsymbol{P} \leftarrow \boldsymbol{P} - {\eta}{\nabla_{\boldsymbol{P}}}L_{\rm CAPT} $ \\
\quad \quad \textbf{end for} \\
\quad \textbf{end for}
\end{tabularx}
\end{flushleft}
\end{table*}

\begin{table*}[t]
\centering
\caption{The average performance for different $\epsilon$ and shots. The \textbf{best} and second best results are highlighted under each metric}
\label{tab:2}



\begin{tabular}
{l lcccccccc>{\columncolor{lightgray!20}}c ccccccc
>{\columncolor{lightgray!20}}c}
 \toprule 

         \multicolumn{2}{c}{} & \multicolumn{2}{c}{\textbf{1 shot}}  & \multicolumn{2}{c}{\textbf{4 shot}} & \multicolumn{2}{c}{\textbf{16 shot}} & \multicolumn{2}{c}{\textbf{All}}& \multicolumn{1}{c}{}  \\

  \cmidrule(lr){3-11}

\textbf{$\epsilon$} &
\textbf{Method}
      & \textbf{Acc.}
        & \textbf{Rob.}
            & \textbf{Acc.}
                & \textbf{Rob.}
                    & \textbf{Acc.}
                        & \textbf{Rob.}
                            & \textbf{Acc.}
                                & \textbf{Rob.}

\tabularnewline
\midrule


\multirow{6}{*}{1/255} & 
HEP  & 45.2  &  32.1  & 45.2&  32.1 &45.2&  32.1&45.2&  32.1
\tabularnewline
 &
AVP &  44.6 &  31.6 & 45.0  &  32.4 &  45.7   &  33.6 &  50.6  &  39.0
\tabularnewline
   &
PAFT  &  30.6  &  21.7   & 46.9 &  34.4 &  66.4   &  \underline{ 51.0} &  71.1   &  56.9
\tabularnewline
 &
APT-UC & \underline{51.3}  &  \textbf{35.1}   & \underline{58.2}  &\underline{40.8}    & 66.5   & 49.0&  70.9  & 54.3
\tabularnewline
 &
APT-CSC&  39.9 & 26.2   & 54.3 &  37.8 & \underline{ 66.6 }   & 49.1 & \underline{73.5 }   &\underline{57.1}   
\tabularnewline
  \cmidrule(lr){2-11} 
 &
CAPT&  \textbf{51.8}  &  \underline{32.6 }   &  \textbf{62.3} &\textbf{42.7 } &  \textbf{72.9  } & \textbf{ 52.7}& \textbf{79.8}    & \textbf{60.1}
\tabularnewline
\midrule

\multirow{6}{*}{4/255} & 

HEP  &  33.0 &  10.3   &  33.0 &  10.3  & 33.0 &  10.3  &  33.0 &  10.3 
\tabularnewline
& 
AVP&  32.2 & 10.5  & 32.4 & 10.8 &  32.7  &  11.3 & 34.4  & 13.1
\tabularnewline
& 
PAFT  & 19.2  & 8.5  & 32.4&  13.9  & 51.5  &  \underline{22.9}  & 54.9   &  \underline{27.5}  
\tabularnewline
& 
 APT-UC&  \underline{ 33.1}  & \underline{11.4 }   &  41.8& \underline{15.2 }    & 51.1  &  20.2 &  54.9   & 24.8
  \tabularnewline
   &
APT-CSC&  28.1  &  8.1  &\underline{41.9}    &  14.4  & \underline{ 54.2 }    &  20.7 & \underline{59.4}      &  27.0
\tabularnewline
\cmidrule(lr){2-11} 
 &
CAPT & \textbf{42.1 }  &\textbf{18.8 }   &\textbf{ 55.1} & \textbf{27.4}  & \textbf{69.1 }  & \textbf{ 38.2 }& \textbf{78.0}  & \textbf{45.2}
\tabularnewline

\bottomrule
\end{tabular}
\end{table*}

\begin{table*}
  \caption{Zero-shot performance. AVP, APT and CAPT were tunned with 100 shots, $\epsilon=4/255$}
  \label{tab:zero-shot}
  \begin{tabular}{c|c|p{1.6cm}<{\centering}p{1.6cm}<{\centering}p{1.6cm}<{\centering}p{1.6cm}<{\centering}p{1.6cm}<{\centering}p{1.6cm}<{\centering}}
    \toprule
        \textbf{Dataset} & \textbf{Metric} & \textbf{TeCoA} & \textbf{HEP} & \textbf{AVP} & \textbf{APT-UC} & \textbf{CAPT \scriptsize w/o CG} & \textbf{CAPT} \\
    \midrule
        \multirow{2}{*}{Average} & Acc. & 32.17 & 33.01 & 32.11 & 33.23 & 49.69 & \textbf{50.27} \\ 
        ~ & Rob. & 10.40 & 10.34 & 11.19 & 12.14 & 17.92 & \textbf{20.11} \\ 
    \midrule
        \multirow{2}{*}{ImageNet} & Acc. & 39.60 & 39.90 & 39.53 & 40.34 & \textbf{66.20} & 65.67 \\ 
        ~ & Rob. & 10.30 & 10.30 & 11.32 & 12.15 & 21.78 & \textbf{24.48} \\ 
    \midrule
        \multirow{2}{*}{FGVCAircraft} & Acc. & 6.50 & 7.00 & 6.50 & 7.10 & 9.69 & \textbf{10.29} \\ 
        ~ & Rob. & 0.40 & 0.50 & 0.60 & 0.80 & 2.43 & \textbf{3.12} \\ 
    \midrule
       \multirow{2}{*}{EuroSAT} & Acc. & 16.40 & \textbf{20.30} & 16.30 & 16.90 & 14.06 & 13.20 \\ 
        ~ & Rob. & 11.00 & 9.20 & \textbf{11.20} & \textbf{11.20} & 3.22 & 1.98 \\ 
    \midrule
        \multirow{2}{*}{Caltech101} & Acc. & 77.40 & 77.40 & 77.30 & 78.30 & 90.14 & \textbf{90.18} \\ 
        ~ & Rob. & 42.80 & 42.80 & 45.10 & 45.70 & 60.65 & \textbf{64.99} \\ 
    \midrule
        \multirow{2}{*}{StanfordCars} & Acc. & 10.30 & 10.30 & 10.40 & 12.10 & 37.28 & \textbf{40.68} \\ 
        ~ & Rob. & 0.90 & 0.90 & 1.10 & 1.50 & 6.68 & \textbf{8.89} \\ 
    \midrule
        \multirow{2}{*}{Food101} & Acc. & 20.30 & 21.70 & 20.20 & 23.90 & \textbf{62.30} & 61.24 \\ 
        ~ & Rob. & 3.10 & 3.20 & 3.40 & 3.80 & 10.74 & \textbf{12.03} \\ 
    \midrule
        \multirow{2}{*}{OxfordPets} & Acc. & 60.90 & 61.50 & 60.90 & 65.20 & 79.69 & \textbf{82.01} \\ 
        ~ & Rob. & 14.60 & 14.30 & 16.40 & 21.90 & 29.35 & \textbf{35.73} \\
    \midrule
        \multirow{2}{*}{OxfordFlowers} & Acc. & 31.40 & 30.50 & 31.30 & 29.10 & 44.94 & \textbf{49.37} \\ 
        ~ & Rob. & 9.20 & 8.80 & 9.80 & 10.00 & 17.45 & \textbf{18.71} \\ 
    \midrule
        \multirow{2}{*}{DTD} & Acc. & 23.70 & 26.30 & 23.40 & 24.30 & \textbf{34.40} & 32.98 \\ 
        ~ & Rob. & 10.40 & 11.60 & 11.00 & 11.60 & 14.54 & \textbf{16.31} \\ 
    \midrule
        \multirow{2}{*}{SUN397} & Acc. & 32.00 & 32.00 & 32.00 & 32.80 & \textbf{54.05} & 53.52 \\ 
        ~ & Rob. & 5.90 & 5.90 & 6.60 & 7.40 & 15.63 & \textbf{18.20} \\ 
    \midrule
        \multirow{2}{*}{UCF101} & Acc. & 35.40 & 36.20 & 35.40 & 35.50 & \textbf{53.79} & \textbf{53.79} \\ 
        ~ & Rob. & 5.80 & 6.20 & 6.60 & 7.50 & 14.67 & \textbf{16.79} \\ 
    \bottomrule
  \end{tabular}
\end{table*}


\bibliographystyle{ACM-Reference-Format}
\bibliography{sample-base}








